\def\year{2018}\relax
\documentclass[letterpaper]{article} 
\usepackage{aaai18}  
\usepackage{times}  
\usepackage{helvet}  
\usepackage{courier}  
\usepackage{url}  
\usepackage{graphicx}  
\usepackage{amsmath}
\usepackage{amssymb}
\usepackage{amsthm}
\usepackage{algorithm}
\usepackage{algorithmic}

\newtheorem{mydef}{Definition}
\newtheorem{thm}{Theorem}
\newtheorem{lma}{Lemma}

\title{Diverse Exploration for Fast and Safe Policy Improvement}

\author{Andrew Cohen \\ Binghamton University \\acohen13@binghamton.edu \And Lei Yu \\ Binghamton University \\ Yantai University \\ lyu@binghamton.edu \And Robert Wright \\ Assured Information Security \\ wrightr@ainfosec.com}

\begin{document}
\maketitle
\begin{abstract}
We study an important yet under-addressed problem of quickly and safely improving policies in online reinforcement learning domains. As its solution, we propose a novel exploration strategy - diverse exploration (DE), which learns and deploys a diverse set of safe policies to explore the environment. We provide DE theory explaining why diversity in behavior policies enables effective exploration without sacrificing exploitation. Our empirical study shows that an online policy improvement algorithm framework implementing the DE strategy can achieve both fast policy improvement and safe online performance.
\end{abstract}

\section{Introduction}
Recent advances in autonomy technology have promoted the widespread emergence of autonomous
agents in various domains such as autonomous vehicles, online marketing, and financial management. Common to all of these domains is the requirement to {\it quickly} improve from the current policy/strategy in use while ensuring {\it safe} operations. We call this requirement the Fast and Safe (policy) Improvement (FSI) problem. On one hand, fast policy improvement demands that an agent acquire a better policy quickly (i.e, through fewer interactions with the environment). On the other hand, a new policy to be deployed needs to be safe - guaranteed to perform at least as well as a baseline policy (e.g., the current policy). Untested policies and/or unrestricted exploratory actions that potentially cause degenerated performance are not acceptable. 

Reinforcement Learning (RL)~\cite{sutton} has great potential in enabling robust autonomous agents that learn and optimize from new experiences. The exploration/exploitation problem is a well studied RL problem. Past research has focused mainly on learning an optimal or near-optimal policy, instead of the FSI problem. Conventional exploration strategies do not provide a suitable trade-off between exploitation and exploration to achieve both fast and safe policy improvement. They make potentially sub-optimal and unsafe exploratory action choices (either blindly like $\epsilon$-greedy~\cite{sutton} or intentionally to reduce uncertainty like R-MAX~\cite{Braf-Tenn03}) at the ``state'' level,  and achieve exploration by ``deviating'' from the best policy according to current knowledge. 

In this work, we take a radically different approach to exploration by performing exploration over the space of stochastic policies. We propose Diverse Exploration (DE) which learns and deploys a {\it diverse} set of {\it safe} policies to explore the environment. Following the insight that in almost all cases, there exist different safe policies with similar performance for complex problems, DE makes exploratory decisions at the ``policy'' level, and achieves exploration at little to no sacrifice to performance by searching policy space and ``exploiting'' multiple diverse policies that are safe according to current knowledge. 
 
The main contributions of this paper are four-fold. 
First, it formally defines the FSI problem. Second, it proposes a new exploration strategy DE as a solution. Third, it provides DE theory which shows how diversity in behavior policies in one iteration promotes diversity in subsequent iterations, enabling effective exploration under uncertainty in the space of safe policies. Finally, it proposes a general algorithmic framework for DE. The framework iteratively learns a diverse set of policies from a single batch of experience data and evaluates their quality through off-policy evaluation by importance sampling before deploying them. We compare this to a baseline algorithm, referred to as SPI (safe policy improvement), which follows the same framework but only learns and deploys a single safe policy at every iteration. Experiments on three domains show that the DE framework can achieve both safe performance and fast policy improvement.

\section{Preliminaries}\label{hcope}
RL problems can be elegantly described within the context
of Markov Decision Processes (MDP)~\cite{puterman2009markov}. An MDP, $M$, is defined as a 5-tuple, $M =(S,A,P,\mathcal{R},\gamma)$, where $S$ is a fully observable finite set of states, $A$ is a finite set of possible actions, $P$ is the state transition model such that $P(s'|s,a) \in [0,1]$ describes the probability of transitioning to state $s'$ after taking action $a$ in state $s$, $\mathcal{R}_{s,s'}^a$ is the expected value of the immediate reward $r$ after taking $a$ in $s$, resulting in $s'$, and $\gamma\in (0,1)$ is the discount factor on future rewards.
A solution to an MDP is a policy, $\pi(a|s)$ which provides the probability of taking action $a$ in state $s$ when following policy $\pi$. The quality or performance of a policy is determined by the expected value that can be obtained by following it from any given state. In RL scenarios $P$ and $\mathcal{R}$ are unknown and $\pi$ must be learned from experiences that take the form of samples. Experience samples are single-step observations of transitions from the domain.  They are represented by tuples, $(s_t,a_t,r_{t+1},s_{t+1})$, which consist of a state $s_{t}$, an action $a_{t}$, the next state $s_{t+1}$, and the immediate reward $r_{t+1}$. A {\it trajectory} of length $T$ is an ordered set of transitions: $\tau= \{s_0,a_0,r_1,s_1,a_1,r_2,...,s_{T-1},a_{T-1},r_{T}\}$.  

It is a challenging problem to estimate the performance of a policy without deploying it. To address this challenge the authors of \cite{ThomasHCOPE} proposed high-confidence off-policy evaluation (HCOPE) methods which lower-bound the performance of a {\it target policy}, $\pi_p$, based on a set of trajectories, $\mathcal{D}$, generated by some behavior policy (or policies), $\pi_q$. In their work, the (normalized and discounted) {\it return} of a trajectory is defined as: $R(\tau)= ((\sum_{t=1}^{T}\gamma^{t-1} r_t)-R_{\_\_})/(R_{\text{+}}-R_{\_\_})\in [0,1]$, where $R_{\text{+}}$ and $R_{\_\_}$ are upper and lower bounds on $\sum_{t=1}^{T}\gamma^{t-1} r_t$.

HCOPE applies importance sampling~\cite{precup2000eligibility} to produce an unbiased estimator of $\rho(\pi_p)$ from a trajectory generated by a behavior policy, $\pi_q$. The estimator is called the importance weighted return, $\hat{\rho}(\pi_p|\tau,\pi_q)$, and is given by: $\hat{\rho}(\pi_p|\tau,\pi_q)=R(\tau)w(\tau,\pi_p,\pi_q)$, where $w(\tau,\pi_p,\pi_q)$ is the importance weight: $w(\tau,\pi_p,\pi_q)=\prod_{t=1}^{T}\frac{\pi_p(a_t|s_t)}{\pi_q(a_t|s_t)}$. Based on a set of importance weighted returns, HCOPE provides a high confidence lower bound for $\rho(\pi_p)$. Let $X_1,...,X_n$ be $n$ random variables, which are independent and all have the same expected value, $\mu=\mathbb{E}[X_i]$. HCOPE considers $\hat{\rho}(\pi_p|\tau_i,\pi_i)$ as $X_i$, and so $\mu=\rho(\pi_p)$.

One difficulty with this approach is importance weighted returns often come from distributions with heavy upper tails, which makes it challenging to estimate confidence intervals based on samples. In~\cite{ThomasHCPI} the authors studied the effectiveness of three methods; concentration inequality, Student's $t$-test, and bootstrap confidence interval. 
We adopt the $t$-test due to its good performance and computational efficiency. Under mild assumptions of the central limit theorem, the distribution of the sample mean approximates a normal distribution, and it is appropriate to use a one-sided Student's $t$-test to get a $1-\delta$ confidence lower bound on the performance of a policy. In~\cite{ThomasHCPI}, policies deemed safe by the $t$-test are called {\it semi}-safe since the estimate is based on possibly false assumptions.

\section{Rationale for Diverse Exploration }\label{sec:rationale}
\subsection{Problem Formulation}
\begin{mydef}
Consider an RL problem with an initial policy, $\pi_0$,
a lower bound, $\rho_{\_\_}$, on policy performance, and a confidence level, $\delta (0<\delta <1/2)$, all specified by a user. Let $\pi_1,...,\pi_d$ be $d (d\ge 1)$ iterations of behavior policies  
and $\rho(\pi_i)$ be the performance (expected return) of $\pi_i$.  {\it Fast and Safe Improvement (FSI)} aims at:
$\max (\rho(\pi_d) - \rho(\pi_0))~\text{subject to~}\forall i=1,...,d,\rho(\pi_i)\ge \rho_{\_\_},~\text{with probability at least~} (1-\delta)~\text{per iteration}$.
\end{mydef}

FSI requires that in each iteration of policy improvement, a behavior policy (the policy that gets deployed) $\pi_i$'s expected return is no worse than a bound $\rho_{\_\_}$, with probability at least $1-\delta$. We call such a policy $\pi_i$ a {\it safe policy}. 
Both $\delta$ and $\rho_{\_\_}$ can be adjusted by the user to specify how much risk is reasonable for the application at hand.  $\rho_{\_\_}$ can be the performance of $\pi_0$ or $\pi_{i-1}$. Furthermore, FSI aims at maximally improving the behavior policy within a limited number of policy improvement iterations. This objective is what distinguishes FSI from the safe policy improvement (SPI) problem that enforces only the safety constraint on behavior policies~\cite{petrik16,ThomasHCPI}. 

To achieve exploration within the safety constraint, one could resort to a stochastic safe policy. However, this is often ineffective for fast improvement because the randomness of the policy and hence the exploratory capacity must be limited in order to achieve good performance. 
Alternatively, we propose DE which strives for behavior diversity and performs exploration in the space of stochastic policies. 

\subsection{Advantage of DE Over SPI Solutions}
DE can be thought of as a generalized version of any solution to the SPI problem. DE learns and deploys a diverse set of safe policies instead of a single safe policy (as is typical in SPI) during each policy improvement iteration. The high confidence policy improvement method in~\cite{ThomasHCPI} is an SPI method that applies HCOPE (reviewed earlier) to provide lower bounds on policy performance. For simplicity, from here on we use SPI to refer to a solution to the SPI problem that uses this safety model. The safety guarantees in HCOPE are the result of importance sampling based estimates. 
A problem with SPI, which has not been previously discussed in the literature, stems from a property of importance sampling: data from a single behavior policy can result in very different variances in the estimates for different candidate policies that SPI evaluates for safety. Specifically, variance will be low for policies that are similar to the behavior policy. Thus, deploying a single behavior policy results in an implicit bias (in the form of a lower variance estimate, and hence a better chance of confirming as a safe policy) towards a particular region of policy space with policies similar to the deployed policy. This does not allow SPI to fully explore the space of policies which may obstruct fast policy improvement. 

To overcome this limitation of SPI and address the FSI challenge, we need to generate sufficient exploration while maintaining safety. Our DE solution achieves exactly this. The DE theory later shows why deploying a population of safe policies achieves better exploration than a single safe policy.
Informally, in the context of HCOPE by importance sampling, when {\it diverse} behavior policies are deployed (i.e., by multiple importance sampling) DE leads to {\it uniformity} among the variances of estimators, which gives an equal chance of passing the safety test to different candidate policies/target distributions. Such uniformity in turn promotes {\it diversity} in the behavior policies in subsequent iterations. While iteratively doing so, DE also maintains the average of the variances of estimators (i.e., maintaining utility of the current data for confirming the next round of candidates). In contrast, SPI deploys only one reliable policy among available ones (i.e., by single importance sampling), and gives a heavily biased chance towards the policy that is most similar to the behavior policy, which leads to a limited update to the data.  These theoretical insights are consistent with our intuition that for a population, diversity promotes diversity, while homogeneity tends to stay homogeneous.

\begin{figure}
\begin{center}
\begin{tabular}{cc}
\includegraphics[width=0.325\linewidth]{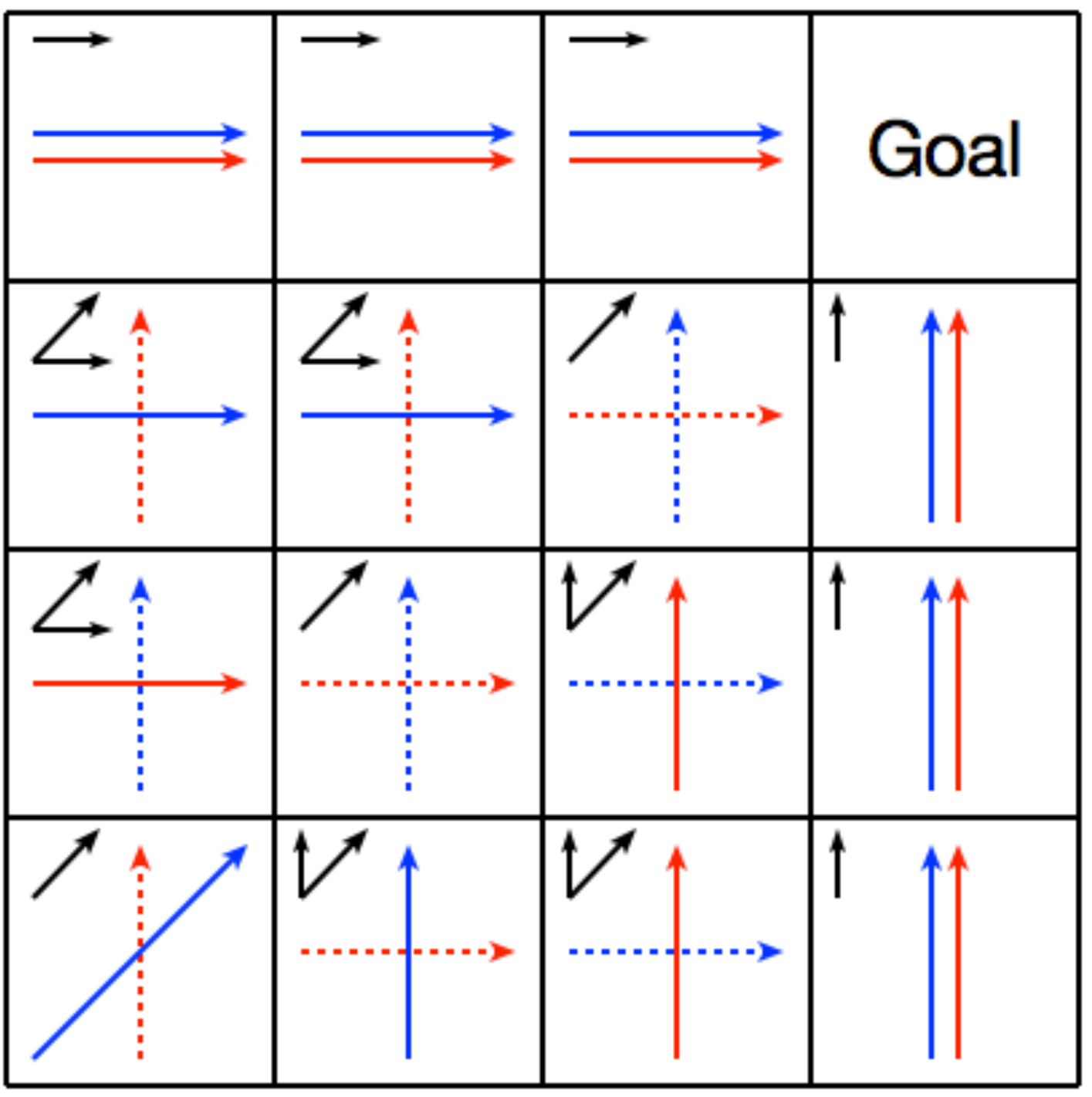}&\includegraphics[width=0.575\linewidth]{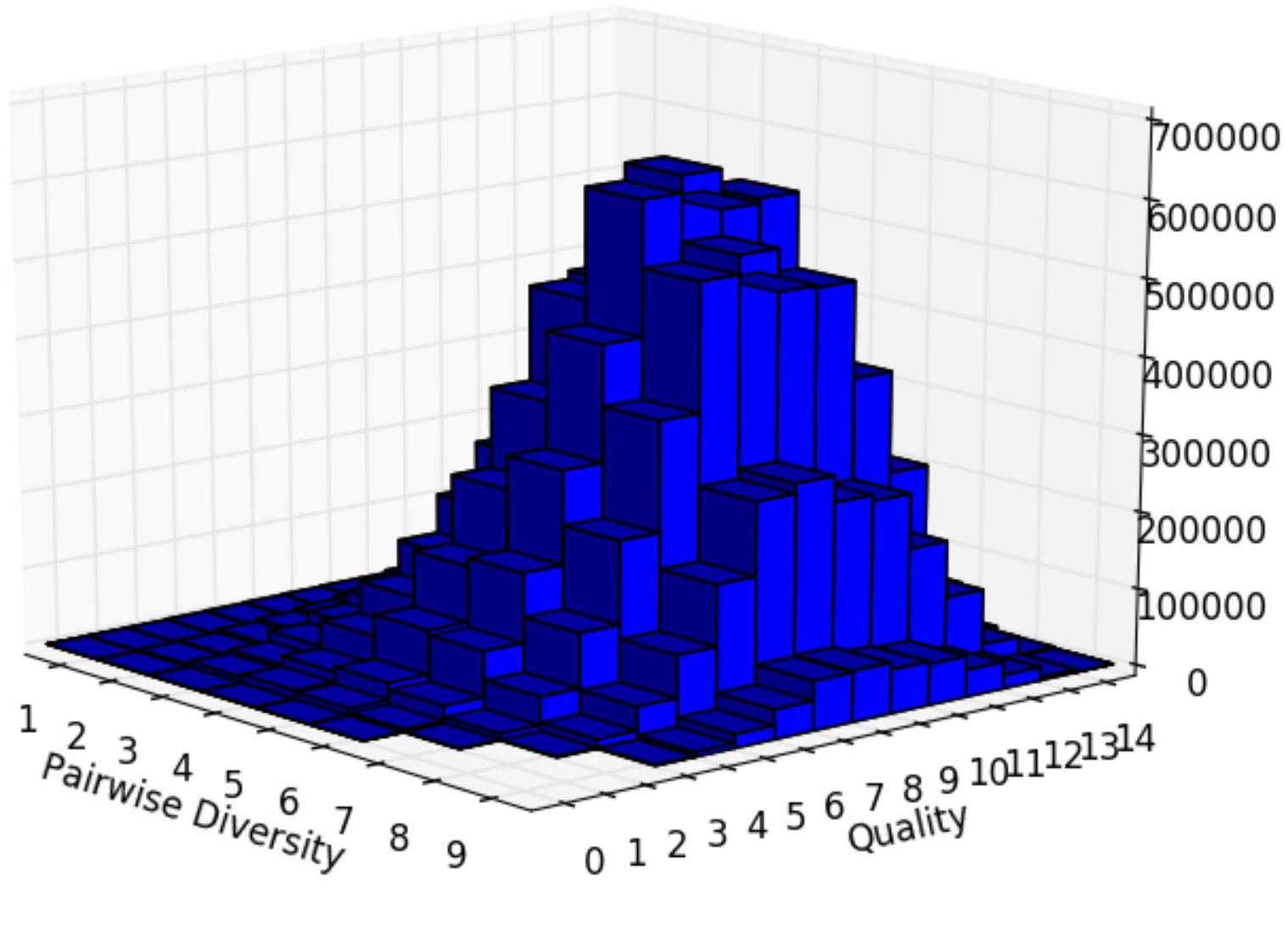}\\
(a)&(b)\\
\end{tabular}
\begin{small}
\caption{A motivating example for : (a)
  A $4\times4$ grid-world with five possible actions ($\nearrow, \uparrow, \rightarrow, \downarrow, \leftarrow$); optimal actions for each state labeled in the upper left corner; an example of two policies (red and blue) of similar quality but different actions at 9 states. (b) A partial view of the distribution of pair-wise diversity (i.e., no. of states two policies differ) across a range of  policy quality (i.e., total extra steps to Goal than optimal over all states).}\label{fig:grid}
\end{small}
\end{center}
\end{figure}

\subsection{Environments with Diverse Safe Policies}
We now show that the behavior diversity needed to realize the synergistic circle of diversity to diversity naturally exists.   
Consider a $4\times4$ grid-world environment in Figure~1 (a). The goal of an agent is to move from the initial (bottom left) state to the terminal (top right) state in the fewest steps. Immediate rewards are always -1. Compared to the standard grid-world problem, we introduce an additional diagonal up-right action to each state that significantly increases the size of the policy search space and also serves to expand and thicken the spectrum of policy quality. From a deterministic point of view, in the standard grid-world, there are a total of $2^9$ optimal policies (which take either up or right in the 9 states outside of the topmost row and rightmost column). All of these policies become sub-optimal at different levels of quality in this extension.

As shown in Figure~1 (a), two policies of similar quality can differ greatly in action choices due to: (1) they take different but equally good actions at the same state; and (2) they take sub-optimal actions at different states. As a result, there exists significant diversity among policies of similar quality within any small window in the spectrum of policy quality. This effect is demonstrated by Figure~1 (b). To manage the space of enumeration, we limit the policies considered in this illustration to the $5^9$ policies that take the diagonal, up, left, down, or right action in the 9 states outside of the topmost row and rightmost column and take the optimal action in other states. The quality of a policy is 
measured in terms of the total {\it extra} steps to Goal starting from each state, compared to the total steps to Goal by an optimal policy. Besides the existence of significant diversity,  another interesting observation from Figure~1 (b) is that as policy quality approaches optimal (extra steps approaches 0), both the total number of policies at a given quality level and the diversity among them decrease. 

In domains with large state and action spaces and complex dynamics, it is reasonable to expect some degree of diversity among safe policies at various levels of quality and the existence of multiple exploratory paths for policy improvement. It is worth noting that in simple domains\footnote{For example, a Markov chain domain with two actions (left or right), and the goal state is at one end of the chain.} where there is significant homogeneity in the solution paths of better policies towards an optimal solution, DE will not be very effective due to limited diversity in sub-optimal policies. In complex domains, the advantage from exploiting diversity among safe policies can also diminish as the quality of safe policies approaches near optimal. Nevertheless, DE will not lose to a safe policy improvement algorithm when there is little diversity to explore, since it will follow the safe algorithm by default. When there is substantial diversity to exploit, our DE theory in the next section formally explains why it is beneficial to do so.

\subsection{Theory on Diverse Exploration}
This section provides justification for how deploying a diverse set of behavior policies, when available, improves uniformity among the variances of policy performance estimates, while maintaining the average of the variances of estimators. This theory section does not address how to effectively identify diverse safe policies.

Importance sampling aims to approximate the expectation of a random variable $X$
with a target density $p(x)$ on $D$ by sampling from a proposal density $q(x)$.
\begin{multline}
\mu = E_p[X] = \int_D f(x)p(x)dx=\int_D f(x)p(x)\frac{q(x)}{q(x)}dx\\
=E_q[\frac{f(x)p(x)}{q(x)}]
\end{multline}

Let $\{p_1,p_2,.\ .,p_r\}$ be a set of $r\ge 2$ target distributions and $\{q_1,q_2,.\ .,q_m\}$ a set of $m\ge 2$ proposal distributions (which correspond to candidate policies, $\pi_p$, and behavior policies, $\pi_q$, in the RL setting, respectively). Note this problem setting is different from traditional single or multiple importance sampling because we consider multiple target distributions ($r\ge 2$). All target and proposal distributions are assumed distinct. For $1 \le j \le r,\ 1 \le t \le m,\ 1\le i \le n,\ n>m$, $X_{j,t,i}=\frac{p_j(x_{i})f(x_{i})}{q_t(x_{i})}$ is the importance sampling estimator for the $j$'th target distribution using the $i$'th sample generated by the $t$'th proposal distribution.

The sample mean of $X_{j,t,i}$ is
\begin{equation}
\mu_{j,t} = \frac{1}{n}\sum_{i=1}^nX_{j,t,i}
\end{equation}

Then, the variance of $\mu_{j,t}$ is
\begin{multline}
var(\mu_{j,t}) = var(\frac{1}{n}\sum_{i=1}^nX_{j,t,i})=\frac{var(X_{j,t,i})}{n}\\
where\ var(X_{j,t,i}) < \infty
\end{multline}

In the context of multiple importance sampling, the sample mean of $X_{j,t,i},\ 1 \le t \le m$ is defined as
\begin{multline}
\mu_{j,{\bf k}} = \frac{1}{n}\sum_{t=1}^m\sum_{i=1}^{k_t}X_{j,t,i}\\
 where\ {\bf k}=(k_1,k_2,.\ .,k_m),\ k_t\ge 0\ \sum_{t=1}^mk_t=n.
\end{multline}
The vector {\bf k} describes how a total of $n$ samples are selected from the $m$ proposal distributions.  $k_t$ is the number of samples drawn from proposal distribution $q_t(x)$.  The second subscript of the estimator $\mu$ has been overloaded
with the vector {\bf k} to indicate that the collection of $n$ samples has been distributed over the $m$ proposal distributions.  There are
special vectors of the form ${\bf k} = (0,.\ .,n,.\ .,0)$ where $k_t=n,\ k_l=0\ \forall\ l\ne t$, which correspond to single importance sampling.  We denote these special vectors as ${\bf k}^{(t)}$ where $1\le t \le m$. When ${\bf k} = {\bf k}^{(t)}$, $\mu_{j,{\bf k}}$ reduces to $\mu_{j,t}$ because all $n$ samples are collected from the $t$'th proposal distribution. 

$\mu_{j,{\bf k}}$ has variance
\begin{align}
var(\mu_{j,{\bf k}})= var(\frac{1}{n}\sum_{t=1}^m\sum_{i=1}^{k_t}X_{j,t,i}) = \frac{1}{n^2}\sum_{t=1}^mk_tvar(X_{j,t,i})
\end{align}
When ${\bf k} = {\bf k}^{(t)}$, $var(\mu_{j,{\bf k}})$ reduces to $var(\mu_{j,t})$.

Given the FSI problem, we are interested in promoting uniformity of variances (i.e., reducing variance of variances) across estimators for {\it an unknown} set of target distributions (candidate policies).  
This brings us to the following constrained optimization problem:
\begin{multline}
{\bf k}^* = arg\ min_{\bf k} \frac{1}{r}\sum_{j=1}^r|var(\mu_{j,{\bf k}}) -\frac{1}{r}\sum_{j=1}^rvar(\mu_{j,{\bf k}})| \label{objf}\\
subject\ to\ {\bf k}^*=(k_1^*,k_2^*,.\ .,k_m^*),\ k_t^*\ge 0\  \sum_{t=1}^mk_t^*=n
\end{multline}
where ${\bf k}^*$ is an optimal way to distribute $n$ samples over $m$ proposal distributions such that the variances of the estimates are most similar (i.e., the average distance between $var(\mu_{j,{\bf k}})$ and their mean be minimized).  If the set of target distributions and the set of proposal distributions are both known in advance, computing ${\bf k}^*$ can be solved analytically. However, in the FSI context, the set of promising candidate target distributions to be estimated and evaluated by a safety test are unknown before the collection of a total of $n$ samples from the set of available proposal distributions which are already confirmed by the safety test in the past policy improvement iteration. Under such uncertainty, it is infeasible to make an optimal decision on the sample size for each available proposal distribution according to the objective function in Equation~(\ref{objf}). Given the objective is convex, the quality of a solution vector {\bf k} depends on its distance to an unknown optimal vector ${\bf k^*}$. The closer the distance, the better uniformity of variances it produces. Lemma~\ref{bound} below provides a tight upper bound on the distance from a given vector {\bf k} to any possible solution to the objective in Equation~(\ref{objf}).

\begin{lma}\label{bound}
Given any vector ${\bf k} = (k_1,k_2,.\ .,k_m)$ such that $k_t\ge 0,\ \sum_{t=1}^m k_t = n$. Let $k_{min} = k_t\ where\ k_t\le k_i\ \forall\ i \ne t$. Then
\begin{multline}
max_{{\bf y}}||{\bf y} - {\bf k}||_{L_1} = 2n-2k_{min}~,\\
where\ {\bf y} = (y_1,y_2,.\ .,y_m),\ y_t\ge 0,\ \sum_{t=1}^m y_t = n.
\end{multline}
\end{lma}

In any given iteration of policy improvement, the SPI approach~\cite{ThomasHCPI} simply picks one of the available proposal distributions and uses it to generate the entire set of $n$ samples. That is, SPI selects with equal probability from the set of special vectors ${\bf k} ^{(t)}$. The effectiveness of SPI with respect to the objective in Equation~(\ref{objf}) depends on the expectation $E[||{\bf k}^{(t)}-{\bf k}^*||_{L_1}]$ where the expectation is taken over the set of special vectors ${\bf k}^{(t)}$ with equal probability. DE, a better, and optimal \textit {under uncertainty of target distributions}, approach based on multiple importance sampling, samples according to the vector ${\bf k}^{DE} = (\frac{n}{m},\frac{n}{m},.\ .,\frac{n}{m})$.

\begin{thm} \label{optimality}
With respect to the objective in Equation~(\ref{objf}), (i) the solution vector ${\bf k}^{DE}$ is worst case optimal; and 
\begin{align}
(ii)\ 0 \le ||{\bf k}^{DE}-{\bf k}^*||_{L_1} \le E[||{\bf k}^{(t)} - {\bf k}^*||_{L_1}] = 2n - 2\frac{n}{m}
\end{align}
where the expectation is over all special vectors ${\bf k}^{(t)}$.
\begin{proof} (Sketch):

(i) $0 \le k_{min} \le \frac{n}{m}$ can be shown by a straightforward pigeonhole argument. In addition, from Lemma~\ref{bound}, smaller $k_{min}$ gives larger upper bound. Since ${\bf k}^{DE}$ has the largest value of $k_{min}=\frac{n}{m}$,   ${\bf k}^{DE}$ is worst case optimal and \begin{align}
0 \le ||{\bf k}^{DE}-{\bf k}^*||_{L_1} \le 2n - 2\frac{n}{m}.
\end{align}

(ii) Follows by evaluating
\begin{multline}
E[||{\bf k}^{(t)} - {\bf k}^*||_{L_1}] = \frac{1}{m}\sum_{t=1}^m ||{\bf k}^{(t)} - {\bf k}^*||_{L_1} = 2n - 2\frac{n}{m}
\end{multline}

\end{proof}
\end{thm}

Theorem~\ref{optimality} part (i) states that the particular multiple importance sampling solution ${\bf k}^{DE}$ which equally allocates samples to the $m$ proposal distributions has the best worse case performance (i.e., the smallest tight upper bound on the distance to an optimal solution). Additionally, any single importance sampling solution ${\bf k}^{(t)}$ has the worst upper bound. Any multiple importance sampling solution vector ${\bf k}$ with $k_t >0~ \forall\ t$ has better worst case performance than ${\bf k}^{(t)}$.  Part (ii) states that the expectation of the distance between single importance sampling solutions and an optimal ${\bf k}^*$ upper bounds the distance between ${\bf k}^{DE}$ and ${\bf k}^*$.  Together, Theorem~1 shows that ${\bf k}^{DE}$ achieves in the worst case optimal uniformity among variances across estimators for a set of $r$ target distributions and greater or equal uniformity with respect to the average case of ${\bf k}^{(t)}$.

\begin{thm} \label{maintains_variance}
The average variance across estimators for the $r$ target distributions produced by ${\bf k}^{DE}$ equals the expected average variance produced by the SPI approach. That is,
\begin{align}
\frac{1}{r}\sum_{j=1}^rvar(\mu_{j,{\bf k}^{DE}})=E[\frac{1}{r}\sum_{j=1}^rvar(\mu_{j,{\bf k}^{(t)}})]
\end{align}
where the expectation is over special vectors ${\bf k}^{(t)}$.
\begin{proof} (Sketch):
It follows from rearranging the following equation:
\begin{align}
\frac{1}{r}\sum_{j=1}^rvar(\mu_{j,{\bf k}^{DE}}) = \frac{1}{r}\sum_{j=1}^r\frac{1}{n^2}\sum_{t=1}^m\frac{n}{m}var(X_{j,t,i})
\end{align}
\end{proof}
\end{thm}

In combination, Theorems \ref{optimality} and \ref{maintains_variance} 
show that DE achieves better uniformity among the variances of the $r$ estimators than SPI while maintaining the average variance of the system. Although DE may not provide an optimal solution, it is a robust approach. Its particular choice of equal allocation of samples is guaranteed to outperform the expected performance of SPI. This leads to the design of our DE algorithm framework in the next section.

\section{Diverse Exploration Algorithm Framework}
Algorithm~\ref{alg1} provides the overall DE framework.
In each policy improvement iteration, it deploys the most recently confirmed set of policies $\mathcal{P}$ to collect $n$ trajectories as uniformly distributed over the $\pi_i \in \mathcal{P}$ as possible. That is, if $|\mathcal{P}|=m$, according to ${\bf k}^{DE} = (\frac{n}{m},.\ .,\frac{n}{m})$. For each trajectory, it maintains a label with the $\pi_i$ which generated it in order to track which policy is the behavior policy for importance sampling later on.
For each set of trajectories $\mathcal{D}_i$ collected from $\pi_i$, partition $\mathcal{D}_i$ and append to $\mathcal{D}_{train}$ and $\mathcal{D}_{test}$ accordingly. Then, from $\mathcal{D}_{train}$ a set of candidate policies is generated in line 8 after which each
is evaluated in line 9 using $\mathcal{D}_{test}$.  If any subset of policies are confirmed they become the new set of policies to deploy in the next iteration. If no new policies are confirmed, the current set of policies are redeployed.  
\begin{algorithm}
\caption{\textsc{DiverseExploration($\pi_{0}$,$r$, $d$, $n$, $\delta$)}}
\label{alg1}
{\bf Input:} $\pi_{0}$: starting policy, $r$: number of candidates to generate, $d$: number of iterations of policy improvement, $n$: number of trajectories to collect per iteration, $\delta$: confidence
	\begin{algorithmic}[1]
		\STATE $\mathcal{P} \gets$ $\{\pi_{0}\}$
		\STATE $\mathcal{D}_{train}$, $\mathcal{D}_{test} =$ $\emptyset$
		\FOR {$j=1~to~d$}
			\FOR {$\pi_i \in \mathcal{P}$}
				\STATE Generate $\frac{n}{|\mathcal{P}|}$ trajectories from $\pi_i$ and append
a fixed portion to $\mathcal{D}_{train}$ and the rest to $\mathcal{D}_{test}$
			\ENDFOR
			\STATE $\rho_{\_\_} =$ t-test$(\mathcal{D}_{test}, \delta, |\mathcal{D}_{test}|)$
			\STATE $\{\pi_1,.\ .,\pi_r\} =$ $GenCandidatePolicies(\mathcal{D}_{train},r)$
			\STATE $passed =$ $EvalPolicies(\{\pi_1,.\ .,\pi_r\} ,\mathcal{D}_{test},\delta,\rho_{\_\_})$
			\IF {$|passed| > 0$}
				\STATE $\mathcal{P} = passed$
			\ENDIF
		\ENDFOR
	\end{algorithmic}
\end{algorithm}

In choosing a lower bound $\rho_{\_\_}$ for each iteration, the $EvalPolicies$ function performs a $t$-test on the normalized returns of $\mathcal{D}_{test}$ without importance sampling. It treats the set of deployed policies as a mixture policy that generated $\mathcal{D}_{test}$. In this way, $\rho_{\_\_}$ reflects the performance of the past policies, and naturally increases per iteration as deployed policies improve and $|\mathcal{D}_{test}|$ increases.

We assume a set of trajectories $\mathcal{D}_{train}$ has been collected by deploying an initial policy $\pi_0$.
The question remains how to learn a set of diverse and good policies which requires a good balance between the diversity and quality of the resulting policies. Inspired by ensemble learning~\cite{Diet01}, our approach learns an ensemble of policy or value functions from $\mathcal{D}_{train}$. The function $GenCandidatePolicies$ can employ any batch RL algorithm such as a direct policy search algorithm as in~\cite{ThomasHCPI} or a fitted value iteration algorithm like Fitted Q-Iteration (FQI)~\cite{ernst2005tree}.  A general procedure for $GenCandidatePolicies$ is given in Algorithm~\ref{alg2}.

\begin{algorithm}
\caption{\textsc{GenCandidatePolicies($\mathcal{D}_{train}$,$r$)}}
\label{alg2}
{\bf Input:} $\mathcal{D}_{train}$: set of training trajectories, $r$: number of candidates to generate

{\bf Output:} set of $r$ candidate policies
	\begin{algorithmic}[1]
		\STATE $\mathcal{C} =$ $\emptyset$
		\STATE $\pi_1 = LearnPolicy(\mathcal{D}_{train})$
		\STATE $\mathcal{C} \gets$ $append(\mathcal{C},\pi_1)$
		\FOR{$i=2~to~r$}
			\STATE $\mathcal{D}' = bootstrap(\mathcal{D}_{train})$
			\STATE $\pi_i = LearnPolicy(\mathcal{D}')$
			\STATE $\mathcal{C} \gets$ $append(\mathcal{C},\pi_i)$
		\ENDFOR
		\STATE \bf{return} $\mathcal{C}$
	\end{algorithmic}
\end{algorithm}

In this paper, we employ a bootstrapping (sampling with replacement) method with an additional subtlety which fits naturally with the fact that trajectories are collected incrementally from different policies. Our intention is to maintain the diversity in the resulting trajectories in each bootstrapped subset of data. With traditional bootstrapping over the entire training set, it is possible to get unlucky and select a batch of trajectories that do not represent policies from each iteration of policy improvement. To avoid this, we bootstrap within trajectories collected per iteration. 
Training on a subset of trajectories from the original training set $\mathcal{D}_{train}$ may sacrifice the quality of the candidate policies for diversity, when the size of $\mathcal{D}_{train}$ is small as at the beginning of policy improvement iterations. Thus, the first policy added to the set of candidate policies is trained on the full $\mathcal{D}_{train}$, and the rest are trained on bootstrapped data.

There is potential for the application of more sophisticated ensemble ideas. For example, one could perform an ensemble selection procedure to maximize diversity in a subset of member policies based on some diversity measure (e.g., pairwise KL divergence between member policies). In this paper, our emphasis is on the feasibility and effectiveness of the basic idea of DE, so we opt for the basic bootstrapping approach.

Although the proposed procedure has some similarity to ensemble learning,
it is distinct in how the individual models are used. Ensemble learning aggregates the ensemble of models into one, while our procedure will validate each derived policy and deploy the confirmed ones independently to explore the environment. As a result, only the experience data from these diverse behavior policies are assembled for the next round of policy learning. 

To validate candidate policies, we need a set of trajectories independent from the trajectories used to generate candidate policies. So, we maintain separate training and test sets $\mathcal{D}_{train}$, $\mathcal{D}_{test}$ by partitioning the trajectories collected from each behavior policy $\pi_i$ based on a
predetermined ratio (1/5, 4/5) and appending to $\mathcal{D}_{train}$ and $\mathcal{D}_{test}$.  $GenCandidatePolicies$ uses only $\mathcal{D}_{train}$ whereas validation in $EvalPolicies$ uses only $\mathcal{D}_{test}$.

Specifically, $EvalPolicies$ uses the HCOPE method (described earlier) to obtain a lower bound $\rho_\_$ on policy performance with confidence $1-\delta$. However, since it performs testing on multiple candidate policies,  
it also applies the Benjamini Hochberg procedure~\cite{benjamini1995} to control the false discovery rate in multiple testing.
A general procedure for $EvalPolicies$ is outlined in Algorithm~\ref{alg3}.

\begin{algorithm}
\caption{\textsc{EvalPolicies($\mathcal{C}$,$\mathcal{D}_{test}$,$\delta$,$\rho$)}}
\label{alg3}
{\bf Input:} $\mathcal{C}$: set of candidate policies, $\mathcal{D}_{test}$: set of test trajectories, $\delta$: confidence, $\rho$: lower bound

{\bf Output:} $passed$: candidates that pass

    \begin{algorithmic}[1]
        \STATE Apply HCOPE t-test $\forall\ \pi_i \in \mathcal{C}$ with $\mathcal{D}_{test},\delta,|\mathcal{D}_{test}|$
        \STATE $passed=$ $\{\pi_i |$ $\pi_i$ deemed safe following FDR control$\}$
        \STATE \bf{return} $passed$
    \end{algorithmic}
\end{algorithm}

\subsection{Relationship to Baseline Algorithm}
Algorithm~\ref{alg3} reduces to the baseline algorithm SPI when the number of candidate policies to generate, $r$, is set to 1.
In this case, $GenCandidatePolicies$ simply returns one policy $\pi_1$ trained on the full trajectory set. The multiple comparison procedure in $EvalCandidatePolicies$
degenerates to a single $t$-test on importance weighted returns. The trajectory collection phase in $DiverseExploration$ becomes a collection of $n$ trajectories from one policy.

In implementation, this baseline algorithm is most similar to the Daedalus2 algorithm proposed in~\cite{ThomasHCPI} (reviewed earlier) with some technical differences. For example, the lower bound $\rho_{\_\_}$ is fixed for each iteration of policy improvement whereas in our algorithm, $\rho_{\_\_}$ increases over iterations.

\section{Empirical Study}
In this section, we present an empirical analysis of DE to evaluate its diversity, safety, and overall effectiveness in on-line learning settings. As a baseline, we use SPI which, like DE, provides a feasible solution to the FSI problem, making a more suitable candidate for comparison than either $\epsilon$-greedy or R-MAX like approaches. Comparing DE with SPI allows us to directly contrast multiple importance sampling vs. single importance sampling.

We use three RL benchmark domains in our analysis: an extended Grid World as described earlier and the classic control domains of Mountain Car and Acrobot \cite{sutton}. To demonstrate the generality of the DE framework we use two markedly different RL algorithms for learning policies. In Grid World we use Covariance Matrix Adaptation, Evolution Strategies (CMA-ES) \cite{hansenCMA}, a gradient-free policy search algorithm that directly maximizes the importance sampled estimate as the objective as in \cite{ThomasHCPI}. In Mountain Car and Acrobot, we use FQI, an off-policy value approximation algorithm, with Fourier basis functions of order 3~\cite{konidaris2011value} for function approximation.  Following~\cite{ThomasHCPI}, we set $\delta=.05$ for all experiments.

Candidate policies are generated as mixed policies, as in ~\cite{ThomasHCPI} and~\cite{Jiang2016}, to control how different a candidate policy can be from a prior behavior policy. A mixed policy $\mu_{\alpha,\pi_0,\pi}$ is defined as a mixture of policies $\pi_0$ and $\pi$ by mixing parameter $\alpha \in [0,1]$: $\mu_{\alpha,\pi_0,\pi}(a|s) := (1-\alpha)\pi(a|s) + \alpha\pi_0(a|s)$. A larger $\alpha$ tends to make policy confirmation easier, at the cost of yielding a more conservative candidate policy and reducing the diversity in the confirmed policies.  In experiments, we use $\alpha=.3$ for Gridworld and  $\alpha=.9$ for Mountain Car/Acrobot. For Mountain Car and Acrobot, we need a high value of $\alpha$ because FQI does not directly maximize the importance sampled estimate objective function as with CMA-ES used for Gridworld. With smaller values of $\alpha$, DE still outperforms SPI but requires significantly more iterations.  

To measure how DE contributes to the diversity of the experiences collected, we use the {\it joint entropy} measure which is calculated over the joint distribution over states and actions. Higher entropy (uncertainty) means higher diversity in experienced (s,a) pairs, which reflects more effective exploration to reduce the uncertainty in the environment.

\begin{figure}
\begin{center}
\begin{tabular}{cc}
\includegraphics[width=0.465\linewidth]{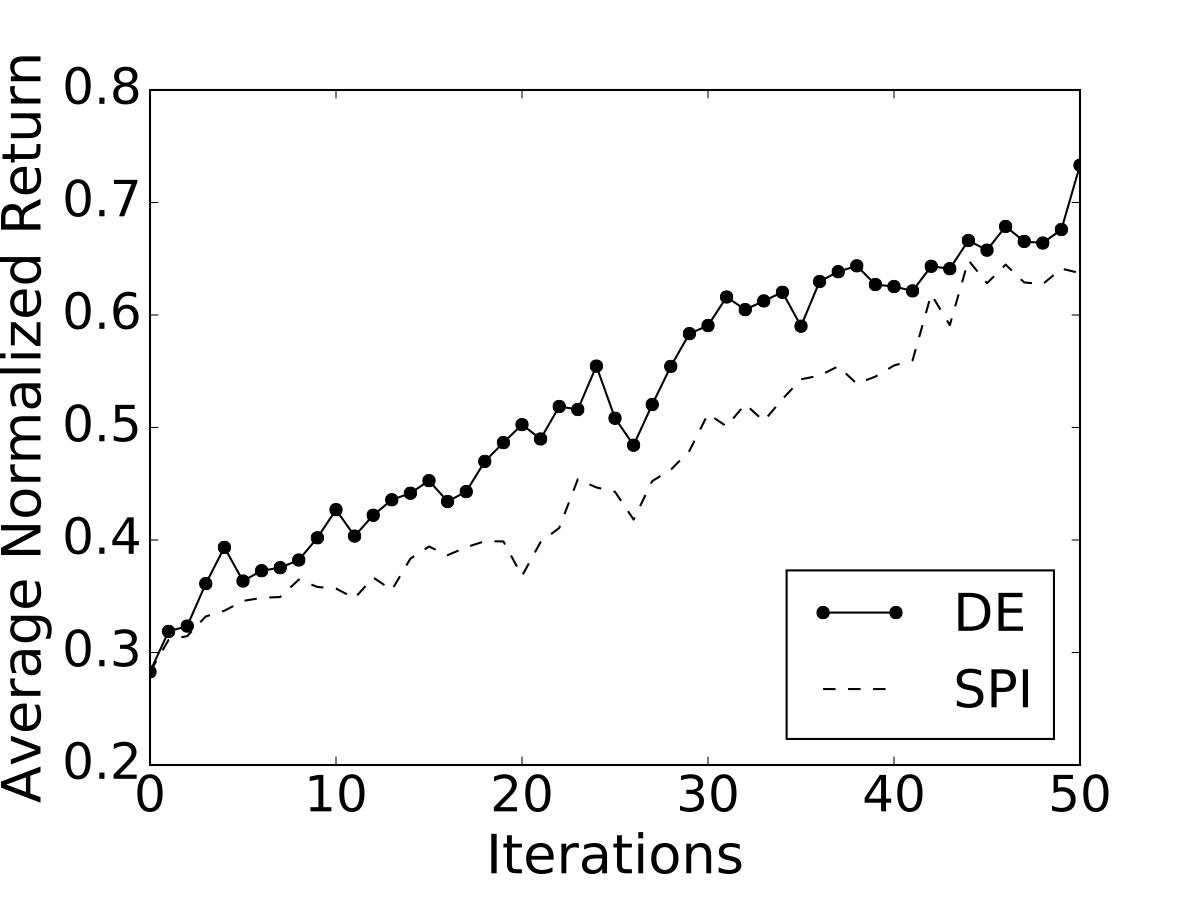} &\hspace{-0.25cm}\includegraphics[width=0.465\linewidth]{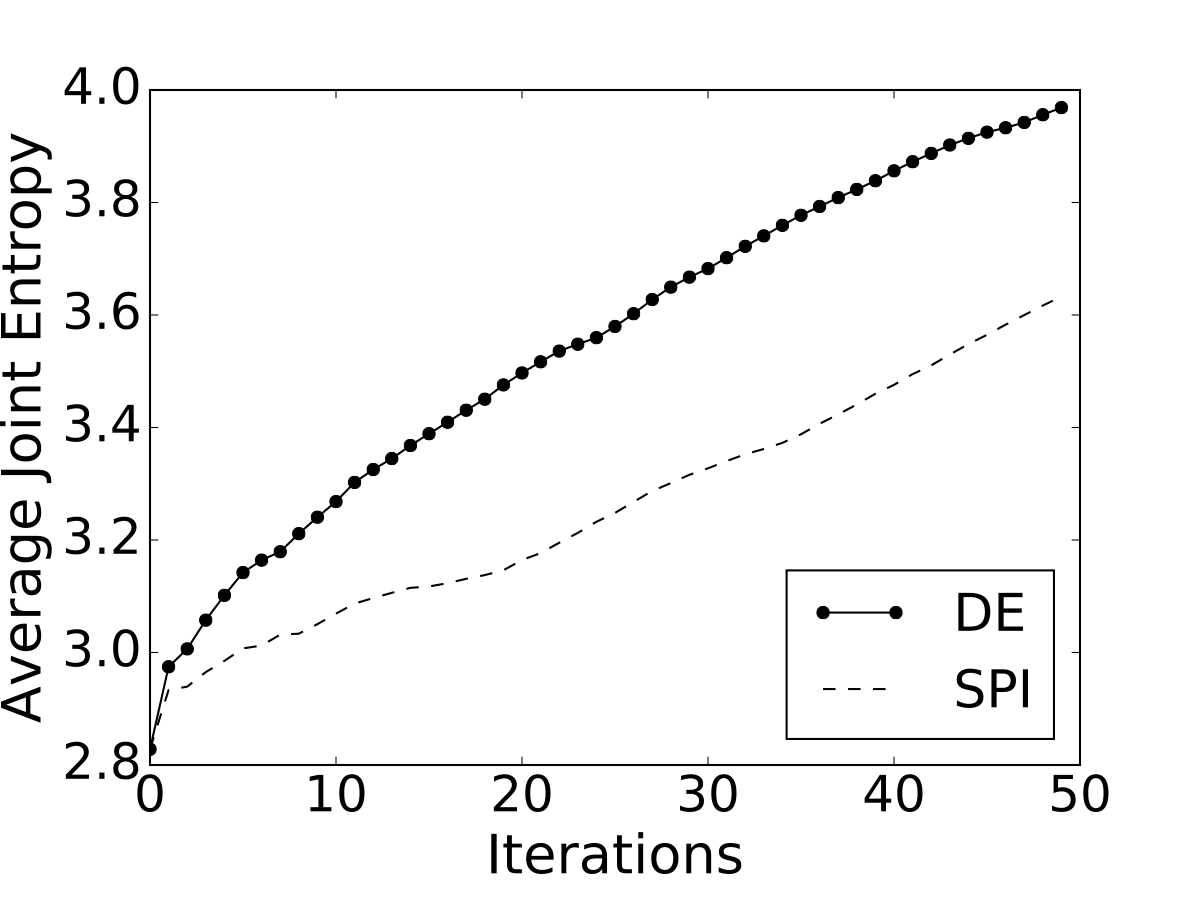}\\
(a) & (b) \\
\end{tabular}
\begin{small}
\caption{(a) Average normalized returns over 50 runs of policy improvement (b) Diversity in experienced $(s,a)$ pairs}
\label{fig:avgruns}
\end{small}
\end{center}
\end{figure}
Figure \ref{fig:avgruns} shows the results comparing DE with SPI on Grid World.  We can see that DE succeeds in our FSI objective of learning more quickly and reliably than SPI does. DE’s deployed policies obtain a higher average return from iteration 7 onward and ultimately achieve a higher return of .73 compared to .65 from SPI. To be clear, each point in the two curves shown in Figure \ref{fig:avgruns} (a) represents the average (over 50 runs) of the {\it average} normalized return of a total of $n=40$ trajectories collected during a policy improvement iteration. To test the significance of the results we ran a two-sided paired $t$-test at each iteration and found $p < .001$. Further, Figure~\ref{fig:avgruns} (b) clearly shows that DE is superior in terms of the joint-entropy of the collected sample distribution, meaning DE collects more diverse samples.  We attribute DE's advantage in overall performance to the significant increase in sample diversity. 

Ideally, an FSI solution will derive and confirm an optimal policy $\pi^*$ in as few iterations as possible, although determining if a given policy is optimal can be difficult in complex domains.  In Grid World, this is not a difficulty as there are 64 distinct optimal policies $\pi^*$.  For these experiments we computed the average number of iterations required to confirm at least one $\pi^*$.  DE achieved this after 16 iterations whereas SPI achieved this after 22 iterations.  This translates to a 240 trajectory difference on average in favor of DE.  Additionally, DE was able to confirm an optimal policy in all 50 runs whereas SPI was unsuccessful in 5 runs. 

For conciseness of presentation, Table \ref{results-tab:1} shows the performance results of the two methods over all three domains in the form of average aggregate normalized return.  This statistic corresponds to the area under the curve for performance curves as shown in Figure~\ref{fig:avgruns} (a).  Higher values indicate faster policy improvement and more effective learning.  The results show that DE succeeds in learning and deploying better performing policies more quickly than SPI.
\begin{table}
\begin{center}

\begin{tabular}{@{}lll@{}}\hline
Domain       & \multicolumn{1}{c}{SPI} & \multicolumn{1}{c}{DE} \\\hline
Grid World   & 604.970                 & \textbf{675.562}       \\
Mountain Car & 362.038                 & \textbf{381.333}       \\
Acrobot      & 417.145                 & \textbf{430.146}       \\ \hline
\end{tabular}
\caption{Average aggregate normalized returns. Bold results are significant improvements ($p \leq .001$)}\label{results-tab:1}
\end{center}
\end{table}

Finally, to evaluate the safety of deployed policies we also compute the empirical error rates (the probability that a policy was incorrectly declared safe). In all experiments the empirical error for DE is well below the 5\% threshold.  Combined these results demonstrate that DE can learn faster and more effectively than SPI without sacrificing safety.

\section{Related Work}
Some recent studies on safe exploration~\cite{Garc-Fern12,Mold-Abbe12,Turchetta2016,achiamCPO,leeCBR} provide safety guarantees during exploration. Their notion of safety is to avoid unsafe states and actions which can cause catastrophic failures in safety critical applications. In contrast, our notion of safety in this paper is defined at the policy level instead of the state and action level. A safe policy must perform at least as well as a baseline policy. A recent work on deep exploration~\cite{osband2016} alluded to a similar idea of exploring the environment through a diverse set of policies, but it does not address the safety issue. Recent advances in approximate policy iteration have produced safe policy improvement methods such as conservative policy iteration~\cite{kakade2002} and its derivatives~\cite{Yadkori2016,Pirotta13}, and off-policy methods~\cite{Jiang2016,petrik16,ThomasHCPI} which decide safe policies based on samples or model estimates from past behavior policies. 
These methods do not perform active exploration during policy improvement.
Manipulating behavior distributions has been explored but with the objective to find an optimal behavior policy to use as a proposal policy for a known target policy~\cite{HannaBPS}.

\section{Conclusions and Future Work}
We have provided a novel exploration strategy as the solution to the FSI problem and the DE theory explaining the advantage of DE over SPI. We have shown that the DE algorithm framework can achieve both safe and fast policy improvement and that it significantly outperforms the baseline SPI algorithm. 

We have only studied some special instances of the DE algorithm under the proposed general framework.  It is natural to incorporate other importance sampling estimators such as~\cite{Jiang2016,ThomasMAGIC,wangSWITCH} into the framework. It would be interesting to see how DE can be integrated with other safe policy improvement algorithms~\cite{petrik16}. Another future direction is to investigate how to optimally generate diverse policies to fully capitalize on the benefit of DE as evidenced in this work.

\subsection*{Acknowledgements} 
The authors would like to thank Xingye Qiao for providing insights and feedback on theorem proofs and the Watson School of Engineering for computing support.
\bibliography{Cohen-Yu_references}
\bibliographystyle{aaai}
\end{document}